# Extracting PICO elements from RCT abstracts using 1-2gram analysis and multitask classification

Xia Yuan[1], Liao xiaoli[1], Li Shilei[2], Shi Qinwen[3], Wu Jinfa[4], Li Ke[*]

**Abstract**

    The core of evidence-based medicine is to read and analyze numerous papers in the medical literature on a specific clinical problem and summarize the authoritative answers to that problem. Currently, to formulate a clear and focused clinical problem, the popular PICO framework is usually adopted, in which each clinical problem is considered to consist of four parts: patient/problem (P), intervention (I), comparison (C) and outcome (O). In this study, we compared several classification models that are commonly used in traditional machine learning. Next, we developed a multitask classification model based on a soft-margin SVM with a specialized feature engineering method that combines 1-2gram analysis with TF-IDF analysis. Finally, we trained and tested several generic models on an open-source data set from BioNLP 2018. The results show that the proposed multitask SVM classification model based on 1-2gram TF-IDF features exhibits the best performance among the tested models.

## 1 Introduction

    Evidence-based medicine (EBM) is a major branch of the medical field. Its purpose is to present statistical analyses of issues of clinical focus based on reading, analyzing, and integrating numerous papers in the medical literature [1]. The PubMed database is one of the most commonly used databases in EBM [2][3]. Successful EBM applications, which rely on abundant research-based evidence combined with clinical expertise [4] and systematic reviews, can effectively assist in clinical decision-making. In most cases, the PICO framework is used to develop a well-defined, focused description clinical problem. In this framework, clinical issues are broken down into four components: patient/problem (P), intervention (I), comparison (C), and outcome (O) [4]. In 2014, Abigail M Methley et al. compared the PICO search method with several other search methods and found that the PICO method can significantly improve the efficiency of a literature search [5]. In 2017, John Rathbone et al. studied the literature screening performed in 10 systematic reviews and found that using the PICO framework can significantly improve the efficiency of literature screening [6].

    The standard EBM analysis process usually consists of the following steps:
1). Use the PICO framework to describe the clinical issue to be studied and develop a literature search strategy based on the formulated problem.
2). In accordance with the developed literature search strategy, attempt to retrieve all documents that meet the stated requirements.
3). Among the retrieved documents, consider the title, abstract, full text and other information to

---

[1] School of Life Science & Technology, University of Electronic Science & Technology of China, bruce_xia6116@163.com

[*] School of Life Science & Technology, University of Electronic Science & Technology of China, colinlike@163.com

Liao Xiaoli and Xia yuan contributed equally.

filter out the articles of interest.

4). Perform a comprehensive analysis of a few of the documents that are ultimately selected and summarize the solutions to and theoretical basis (evidence) for the corresponding clinical problem.

Unfortunately, because the PICO elements are not explicitly identified in the structured abstracts of most medical papers, the retrieval and screening of documents are extremely time-consuming tasks in this era of information explosion. It is often necessary for researchers to thoroughly read each abstract to extract the corresponding PICO information before filtering. Therefore, the ability to automatically extract the PICO elements from the structured abstracts found in PubMed by means of machine learning methods would facilitate the EBM process [7].

In this study, we present a term frequency-inverse document frequency (TF-IDF)-based feature engineering method that incorporates a model based on 1-2gram. We also propose a soft-margin support vector machine (SVM) [8][9] model based on multitask classification for automatically extracting sentence-level PICO elements from structured abstracts in the biomedical literature. The main contributions of this paper are as follows:

1). First, we analyzed the vocabulary and grammatical features of medical English from the linguistic perspective. We also performed a statistical analysis of word frequency on the PICO sentences in our data set. On this basis, we developed a feature construction method that combines 1-2gram with TF-IDF analysis.

2). By designing 6 sets of controlled experiments, we demonstrated the efficiency of the 1-2gram model. We also performed a performance comparison between the TF-IDF feature engineering method and the word2vec word embedding method.

3). We also compared our model with two classic classification methods used in integration learning, i.e., the random forest (RF) method and XGBoost, using the same open-source data set for training and testing.

4). Using the same evaluation indicators, we compared our model with the best two models Naïve Bayes (NB) [3] and long short-time memory (LSTM) [4] from previous studies. Although we used the traditional TF-IDF approach for feature construction, whereas the LSTM model uses word2vec for feature engineering, our model produces better experimental results than the LSTM model does.

## 2 Related Works

Dina Demner-Fushman et al., 2006, studied a method of automatically identifying related information in medical texts [12]. By comparing a series of methods and basic classifiers, including the NB method, the linear SVM method, the decision tree method, a rule-based classifier, an n-gram-based classifier, a position classifier, a document length classifier, a semantic classifier and others, [12] based on a collection of 592 MEDLINE citations, they ultimately found that an automated system that combines domain knowledge with modern statistical methods can help to efficiently extract statements about specific outcomes in medical texts at the micro level.

The first study on the automatic detection and extraction of PICO elements from structured abstracts in the biomedical literature was conducted in 2007 by Demner-Fushman et al. [13]. These authors proposed a rule-based pattern matching method for detecting PICO elements in document abstracts by applying corresponding rules formulated by experts. UMLS, MetaMap, and SemRep tools were used to assist in the text processing necessary to extract biomedical concepts and their interrelationships at the sentence level. On a smaller data set, the method achieved 80% accuracy for patient/problem elements, 86% accuracy for intervention elements and 68%-95% accuracy for outcome elements. Although the research of Demner-Fushman et al. achieved excellent results, this method of pattern matching based on expert annotation has a fatal shortcoming, namely, the requirement for cumbersome manual labeling, which limits the size of the data set [14]. In addition, there may be differences between the annotator and the information seeker in terms of analysis or emphasis [13][15][16]. In the ensuing 10 years, great progress has been made in the related research on this task.

In recent years, the "C" category has often been incorporated into the "I" category because "comparative" elements can refer to other interventions or to the decision not to participate in clinical randomized controlled trials (RCTs) [4], which should be addressed in the intervention category. In fact, very few abstracts with comparison labels are found in PubMed. Moreover, in most PICO studies, C and I elements are merged into the same category in practice because they are considered to form one semantic group [17] [18] [19].

Grace Yuet-Chee Chung et al. 2009 [20] proposed an automated text mining method for automatically identifying intervention elements in abstracts in the RCT literature. These authors analyzed the structure of biomedical texts from the perspective of medical linguistics and summarized the vocabulary and syntactic patterns of intervention-related sentences from numerous RCT abstracts. Then, they proposed a form of coordination construction based on the combination of the identified lexical features and syntactic structures. After training and testing on 203 structured abstracts containing intervention-related sentences, they found that their proposed method achieved high accuracy in the recognition of statements concerning drug treatment interventions.

Chung 2009 [21] studied a method of detecting key P/I/O sentences in RCT abstracts using conditional random fields (CRFs). The method achieved the best F1 values for 38 manually annotated test sums, with values of 83% for I elements and 84% for O elements. Chung's study laid the procedural foundations for the use of structured abstracts in building training corpora that include goals, interventions, participants, outcome measures, methods, results, and conclusions. The corpora used in most research performed after their study have been built following this method.

Florian Boudin and Jian-Yun Nie et al. 2010 [17] proposed a combination of multiple classifier models (Multi-Layer Perceptron, MLP) using weighted linear combinations of predicted scores. They also created a relatively large-scale corpus from 260,000 clinical trials and RCT abstracts, but this corpus was not made public. Their model achieved F1 values of 86.3% for P elements, 67% for I elements, and 56.6% for O elements. Ke-Chun Huang and Charles Chih-Ho Liu et al. [22] proposed an NB model using the top-frequency term, which achieved F1 values of 91% for P elements, 75% for I elements and 88% for O elements.

Di Jin et al. 2018 [11] proposed a classification model based on an LSTM neural network for the automatic extraction of PICO elements from structured abstracts in PubMed. Coupled with the word2vec feature engineering method, the model achieved F1 values of 85.6% for P elements, 78.1% for I elements, and 83.8% for O elements. These authors also constructed a relatively large-scale, fairly complete open-source data set from 489,026 structured abstracts. This data set contains a total of 24,668 structured abstracts containing P/I/O tags and 319,968 sentences with 7 unique tags.

The research studies and methods related to this problem are not restricted to those mentioned above. The work of Florian Boudin 2010 [17] references a variety of methods, such as the RF, SVM, and NB methods. The studies prior to the Di Jin 2018 study were essentially based on traditional machine learning methods. With regard to feature engineering, the vast majority of previous studies used traditional rule-based or word-frequency probability-based statistics. Di Jin 2018 was the first study in which a deep learning method was proposed for extracting P/I/O elements from PubMed abstracts, and that study was also the first time that the word2vec model was used for feature construction. Although some researchers have used the TF-IDF approach for feature construction in previous studies (such as Ke-Chun Huang et al. 2011 [10]), n-gram analysis has not been adopted. Moreover, the potential advantages of n-gram models have not been mentioned in any previous related biomedical texts.

We performed a statistical word-frequency analysis on all PICO sentences in our data set and sorted the results in order of high frequency to low frequency. The results are presented in Table 5. The 10 words with the highest frequency in P sentences are words related to descriptions of the experimental participants (such as patients, women, years, and age). The 10 words with the highest frequency in I sentences are words used to describe interventions (such as group, mg, receive, and placebo). The ten words with the highest frequency in O sentences are words related to the description of results (such as outcome, primary, scale, and measured). In this paper, we propose a TF-IDF-based feature engineering method that combines this word-frequency analysis with a 1-2gram analysis to account for the particularities of the language used in medical texts.

We compared our novel feature engineering method with that of word2vec using the same data set and the same soft-margin SVM modeling approach. We found that the proposed 1-2gram TF-IDF feature engineering method obviously outperforms word2vec.

## 3 Materials and Methods

MEDLINE is an open-access database of medical articles. By June 28, 2012, there were 21,906,254 articles indexed in this database. Searching for and exporting RCT abstracts using search criteria such as MeSH tags would be a lengthy and cumbersome process. To reduce this cumbersome workload, the data used in this study were processed from an open-source data set from BioNLP 2018. The flow chart for this study is shown in Fig 1. The data processing was mainly performed by means of regular matching. For feature engineering, we primarily relied on our 1-2gram TF-IDF model. Finally, we constructed three binary classification models using a standard soft-margin SVM approach for the classification of PICO sentences.

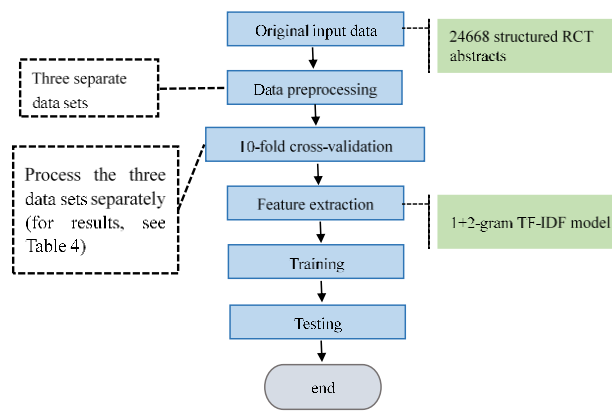

Fig 1. Flow chart of the soft-margin SVM classifier training and testing process. After data preprocessing, three separate data sets were used to train three binary classifier models. Ten-fold cross-validation was applied.

### 3.1 Data sources

The data set used in this study was taken from the proceedings of the BioNLP 2018 workshop (Di Jin et al., 2018) [11]. The data set contains 24,668 PubMed article abstracts with P/I/O tags; 21,198 of the abstracts have P tags, 13,712 have I tags, and 20,473 have O tags. Since these data derived from structured abstract data from PubMed, these structured abstracts mainly include 7 types of tags: participants (P), intervention (I), outcome (O), aim (A), methods (M), results (R), and conclusions (C). Therefore, only these 7 common tags were retained in the data set, and a few other tags were converted into tags similar to one of these 7. In this data, the original 24,668 structured abstracts are also divided into individual sentences. At the same time, each sentence was labeled with its corresponding tag. Some data are shown in the following Table 1.

Table 1: A typical example of an abstract with paragraph headings and the corresponding annotation labels. The PMID of this abstract is 28074672. The heading column shows the structured tag from the original abstract. The sentence column shows the corresponding sentence, and the label column shows the tag we define for our own use.

| Heading | Label | Sentence |
| --- | --- | --- |
| Objective | A | To assess the feasibility of conducting a randomized controlled trial to determine the effectiveness of a twenty-week power… |
| Design | M | Pilot randomised controlled trial. |
| Setting | M | A large-scale twenty-four-hour residential facility in the Netherlands. |
| Subjects | P | Thirty-seven persons with profound intellectual and multiple disabilities. |
| Intervention | I | Participants in the intervention group received a power-assisted exercise intervention… |
| Intervention | I | Participants in the control group received care as usual. |
| Main Measures | O | Trial feasibility by recruitment process and outcomes completion rates… |
| Results | R | Thirty-seven participants were recruited (M age = 32.1)… |
| Results | R | Programme compliance rates ranged from 54.2% to 97.7% with a mean (SD) of 81.5% (13.4). |
| Results | R | Oxygen saturation significantly increased in the intervention group. |
| Results | R | Standardised effect sizes on the difference between groups in outcome varied between 0.02 and 0.62. |
| Conclusions | C | The power-assisted exercise intervention and the trial design were feasible and acceptable to people… |
| Conclusions | C | This pilot study suggests that the intervention improves oxygen saturation… |

## 3.2 Data processing

The original data consist of a total of 24,668 abstracts and 319,968 sentences, of which 21,198 of the abstracts and 27,696 of the sentences have P tags. The specific statistics are shown in Table 2.

Table 2: Summary of the numbers and distributions of abstracts and sentences with P/I/O tags.

| Label | Articles | Sentences |
|---|---|---|
| P | 21198 | 27696 |
| I | 13712 | 24603 |
| O | 20473 | 32526 |

We designed three separate binary classification models for P, I and O classification to mitigate the possible effects of the imbalanced data distribution on the model classification results [23][24] and to avoid sentences with fuzzy labels being assigned as negative samples of non-P/I/O labels, which could affect the feature learning capabilities of the model [22]. First, we filtered out all sentences with non-P/I/O tags from the data set. Then, for the P classification model, we replaced all P tags in the data set with a value of 1 and all non-P tags with a value of 0; for the I classification model, we replaced all I tags in the data set with a value of 1 and all non-I tags with a value of 0; and for the O classification model, we replaced all O tags with a value of 1 and all non-O tags with a value of 0. The data distributions thus obtained are summarized in Table 3 below.

Table 3: Distributions of P/I/O sentences and non-P/I/O sentences in the processed data sets.

| Label | 1 | 0 |
|---|---|---|
| P | 27696 | 57129 |
| I | 24603 | 60222 |
| O | 32526 | 52299 |

Note: 1 represents sentences with the specified label, and 0 represents sentences whose labels do not match the specified label.

We used the stratified sampling method to divide each of the data sets obtained as described above into a training set, a test set and a verification set at a ratio of 8:1:1. For example, for the P classification model, there were 22347 P sentences and 46,095 non-P sentences in the training set, 2723 P sentences and 5550 non-P sentences in the test set, and 2626 P sentences and 5484 non-P sentences in the verification set. The exact distributions are shown in Table 4 below.

Table 4: Distributions of P/I/O sentences and non-P/I/O sentences in the training, test, and verification sets.

| Label | Train | | Test | | Dev | |
|---|---|---|---|---|---|---|
| | 1 | 0 | 1 | 0 | 1 | 0 |
| P | 22347 | 46095 | 2723 | 5550 | 2626 | 5484 |
| I | 19865 | 48557 | 2231 | 5942 | 2507 | 5723 |
| O | 26230 | 42212 | 3219 | 5054 | 3077 | 5033 |

Note: Three data sets were constructed for each model; for example, the values in the P row and the Train column represent the training set for the P classification model, the values in the P row and the Test column represent the test set for the P classification model, and the values in the P row and the Dev column represent the verification set for the P classification model.

## 3.3 Feature extraction

Through a statistical analysis of the word frequencies in the PICO sentences in our data set (the results are shown in Table 5, we found that vocabulary used to describe the experimental participants (such as patients, women, years, and age) usually appears in P sentences, vocabulary related to clinical interventions (e.g., group, mg, received, and placebo) usually appears in I sentences, and vocabulary related to the description of results (e.g., outcome, primary, scale, measured, etc.) usually appears in O sentences. Consequently, these words (terms) are considered as features of the different sentence types.

Table 5: Word-frequency statistics for PICO sentences.

| Sentence type | Top 10 words |
| --- | --- |
| P sentences | Patients, years, women, age, group, study, aged, total, hundred, mean |
| I sentences | Group, patients, mg, received, placebo, weeks, treatment, control, intervention, daily |
| O sentences | Outcome, primary, scale, measured, months, pain, outcomes, treatment, secondary, assessed |

On this basis, we calculated the TF-IDF for each word in every sentence in the data set and then constructed a dictionary, which we used to vectorize each sentence.

$$tf_{i,j} = \frac{n_{i,j}}{\sum_k n_{k,j}} \qquad (1)$$

The term frequency (TF) refers to the frequency with which a given word appears in a sentence. $n_{i,j}$ is the number of occurrences of word $w_i$ in sentence $s_j$, and the denominator is the sum of the occurrences of all words in sentence $s_j$.

$$idf_i = \log \frac{|D|}{|\{j:w_i \in s_j\}|} \qquad (2)$$

where $|D|$ is the total number of sentences in the data set and $|j:w_i \in s_j|$ is the total number of sentences containing the word $w_i$. If this word is not present in the corpus, the denominator in the above expression will be zero; therefore, we add 1 to the denominator, i.e., $|j:w_i \in s_j| + 1$, to prevent division by zero. Finally, the TF-IDF of word $w_i$ in sentence $s_j$ is expressed as

$$tfidf_{i,j} = tf_{i,j} \times idf_i \qquad (3)$$

Medical language is a distinct language [25]. In medical English, especially in medical texts, numerous structures, such as nouns, action nouns, and action noun phrases, are used. Nouns and action nouns mostly consist of single specialized words, whereas an action noun phrase is typically composed of two or more words, and a noun phrase is typically an action noun phrase containing a preposition [26]. Therefore, we added a 1-2gram model to our TF-IDF model. A 1-2gram model was chosen because in English, most prepositions are stop words, and most noun phrases and action noun phrases are no longer than 2 words after the removal of stop words. To verify the property of the 1-2gram model in medical language, we also designed six sets of controlled trials: with all model parameters being the same, we set the n-gram range to 1, 2, 3, 1-2, 1-3, and 2-3, and each of the resulting models was trained and tested on the same data set. The results showed that the property of model is optimal when the n-gram range is 1-2.

## 3.4 Soft-margin SVM

In this study, we replaced the traditional multi-classifier model with three binary classification models to complete the classification task. In terms of the selection of binary classification model, we chose the currently mature SVM classifier. Although text categorization is a typical nonlinear classification problem, we ultimately choose to use SVM classifiers for binary classification after comparing several models. As shown by the results of the previous word-frequency analysis, P and I sentences exhibit a certain overlap in their word-frequency distributions (many P and I sentences contain the same words, such as patients, group, and study). Therefore, after vectorization, the vectors of P and I sentences show a certain linear indivisibility in the vector space; consequently, to increase the generalization ability of the model, we used a linear kernel and a soft margin to train a nonlinear SVM. The soft-margin SVM approach represents an improvement over the standard SVM approach. In a standard SVM, the

model constraint is

$$y_i(w^T x_i + b) \geq 1 \quad , i = 1, \ldots, n \quad (4)$$

where $x_i$ is the vector representation of sentence i and $y_i$ is the label of sentence i, and the objective function is

$$\min \frac{1}{2}\|w\|^2 \quad (5)$$

Linear indivisibility means that for some sample points $(x_i, y_i)$, the constraint that the function interval must be greater than or equal to 1 is not satisfied. Therefore, for each sample point, we introduce a slack variable $\xi_i \geq 0$ such that the function interval plus the slack variable will be greater than or equal to 1; then, the constraint becomes

$$y_i(w^T x_i + b) \geq 1 - \xi_i \quad , i = 1, \ldots, n \quad (6)$$

For each slack variable $\xi_i$, some penalty must be paid; therefore, the objective function becomes

$$\min \frac{1}{2}\|w\|^2 + C \sum_{i=1}^{n} \xi_i \quad (7)$$

Here, C>0 is a penalty parameter, which is utilized to control the relative weight between the two terms in the objective function (which serve to "find the hyperplane with the largest margin" and "minimize the deviation of the data points", respectively). $\xi$ is a variable that needs to be optimized during the model training process. A larger C value corresponds to a greater penalty for misclassification, and a smaller C value corresponds to a smaller misclassification penalty. Therefore, if C is too large, the model can easily be overfitted, whereas if C is too low, the model can easily be underfitted. To choose the appropriate value of the penalty parameter C, we conducted a comparative experiment using values between C=0.1 and C=3.0. We gradually increased the value of C, retraining and testing the model at various intervals. We found that for P and I elements, the model achieved the best F1 values with C=1.0, whereas for O elements, the model achieved the best F1 value with C=0.6. Therefore, we set C=1.0 for P and I classification and C=0.6 for O classification.

## 4 Results

To prove the optimal performance of the 1-2gram model, we designed 6 sets of control experiments. The n-gram range in the TF-IDF analysis was set to 1, 2, 3, 1-2, 1-3, and 2-3; tests were performed using SVM models trained with these n-gram ranges with all other parameters being the same, and the resulting accuracy rate (acc) and F1 values were recorded as evaluation indicators. The test results are presented in the following Table 6. The test results show that the grammar model with 1 to 2 elements exhibits the best performance in terms of both the acc and F1 values. The test results also show that the acc and F1 values are quite similar for n-gram ranges of 1-2 and 1-3. This is because the n-gram range of 1-3 itself contains the n-gram range of 1-2. Therefore, these control experiments confirm the efficiency of the 1-2gram model for the language used in medical texts.

Table 6: Results of n-gram control tests

| n-gram range | P_SVM | | I_SVM | | O_SVM | |
| --- | --- | --- | --- | --- | --- | --- |
| | acc | F1 | acc | F1 | acc | F1 |
| 1 | 0.914 | 0.863 | 0.893 | 0.802 | 0.911 | 0.886 |
| 2 | 0.892 | 0.818 | 0.863 | 0.726 | 0.874 | 0.831 |
| 3 | 0.809 | 0.615 | 0.796 | 0.483 | 0.773 | 0.614 |
| 1-2 | 0.9243 | 0.8792 | 0.8992 | 0.8144 | 0.9154 | 0.8913 |
| 1-3 | 0.9237 | 0.8788 | 0.8983 | 0.8139 | 0.9148 | 0.8907 |
| 2-3 | 0.893 | 0.821 | 0.861 | 0.721 | 0.875 | 0.832 |

Note: The P_SVM column represents the SVM binary classification model designed for P sentences, and the n-gram range column specifies the parameter(s) of the n-gram model we considered in the TF-IDF analysis, where an n-gram range of 1 corresponds to unigrams and an n-gram range of 2 corresponds to bigrams. We present the acc and F1 values obtained through 10-fold cross-validation as evaluation indicators for comparison. Since the results for n-gram ranges of 1-2 and 1-3 differ by less than 0.1, these results are presented up to four significant digits after the decimal point to allow them to be distinguished.

For the six control groups represented in the table, all of the same conditions were used except for the n-gram range, including the data set and all other model parameters.

To test the efficiency of the proposed 1-2gram TF-IDF feature engineering method, we compared it with the word2vec method. First, we obtained summary data for 200,000 RCT articles from Ji Young Lee 2017 [27]. Combined with our existing 24,668 RCT abstracts, we obtained a total of 224,668 abstracts. After a series of processing steps, such as word segmentation and stop word removal, the open-source toolkit gensim was used to train 200-dimensional word vectors. Then, we use the trained word2vec vectors to vectorize all sentences in the data set, performed training and testing using the same soft-margin SVM model, and compared the results with those of the model trained with our 1-2gram TF-IDF features, as presented in Table 7.

Table 7: Results of the TF-IDF and word2vec comparison experiment

|          | P elements |       |       | I elements |       |       | O elements |       |       |
|----------|-----------|-------|-------|-----------|-------|-------|-----------|-------|-------|
|          | P | R | F1 | P | R | F1 | P | R | F1 |
| TF-IDF   | 0.925 | 0.838 | 0.879 | 0.842 | 0.789 | 0.814 | 0.886 | 0.897 | 0.891 |
| word2vec | 0.894 | 0.796 | 0.842 | 0.808 | 0.731 | 0.768 | 0.866 | 0.855 | 0.861 |

Note: The experimental results for the TF-IDF method and the word2vec method compared in this study were obtained using the same soft-margin SVM classification model; the P elements column represents the results for a classification model constructed for P sentences, and the I elements column represents the results for a classification model constructed for I sentences.

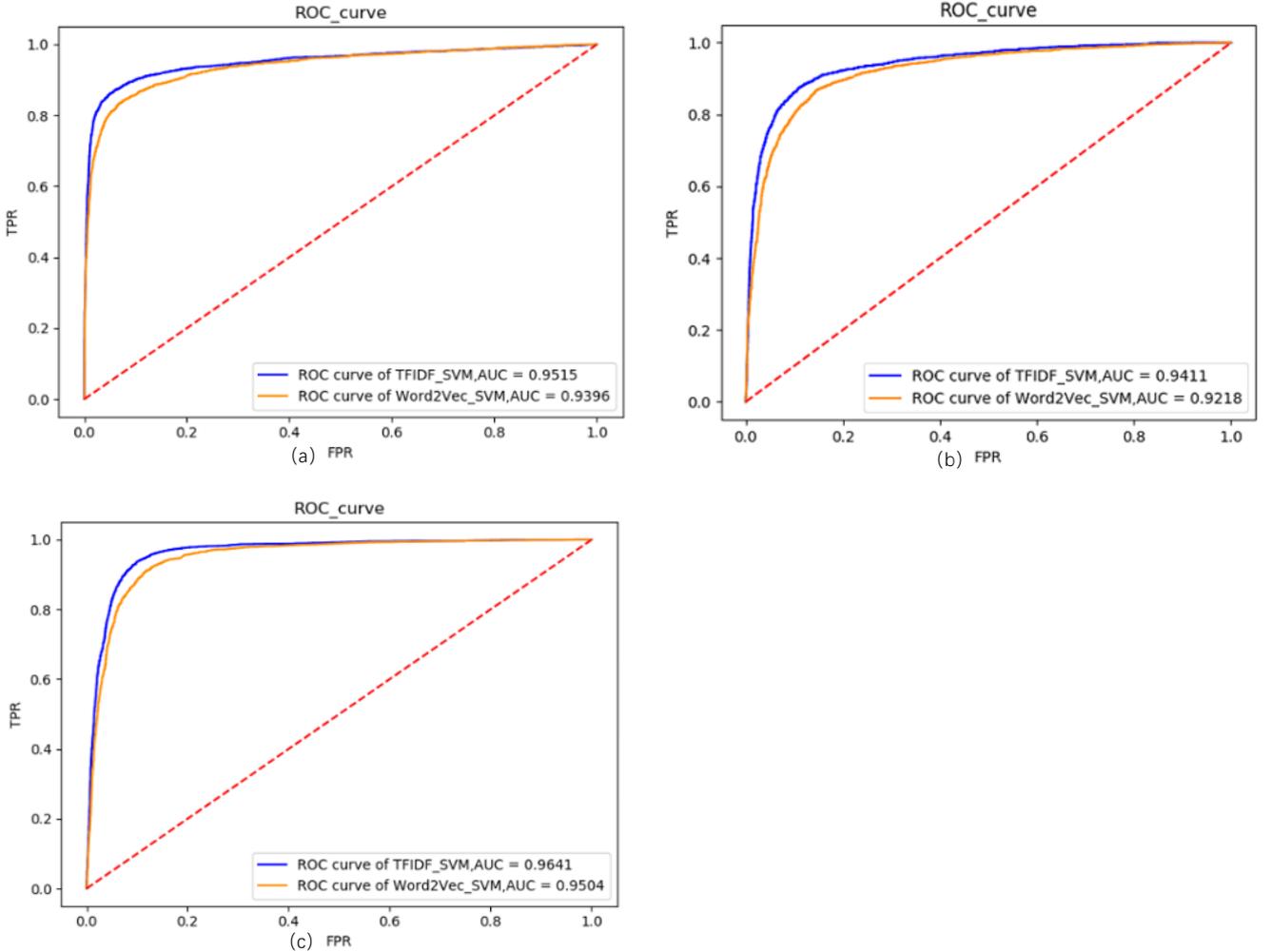

Fig 2. The ROC curves of the TF-IDF and word2vec comparison experiment: (a) results for P elements, (b) results for I

elements, and (c) results for O elements.

It is not difficult to see from Table 7 and Fig 2 that the results (in terms of the P, R, and F1 values) obtained with the TF-IDF model are better than those obtained with the word2vec model. As seen from the results in this table, the differences between the P and R values for the P and I classification models are large, while the differences between the P and R values for the O classification model are relatively small. The reason for this behavior is the overlap in the word-frequency distributions of P and I sentences, which is one of the main factors affecting the R values of the models.

We also compared our method with the RF and XGBoost methods, which are commonly used for SVM modeling and integrated learning, on the same data set. The experimental P, R and F1 values are shown in Table 8. The corresponding ROC curves are given in Fig 3. The experimental results show that the soft-margin SVM model exhibits the best performance on the same data set.

As seen from the results in Table 8, the P values of the SVM model are much higher than those of the XGBoost model for P, I and O elements, but for P and I elements, the R values of the SVM model are lower than those of the XGBoost model. The main reason for this finding is that, as mentioned above, the overlap in the word-frequency distributions of P and I sentences causes these sentences to exhibit strong linear indivisibility in the vector space. Although we introduced the slack variable and the penalty coefficient C into the SVM model, we still must choose the C value appropriately to avoid overfitting. Therefore, the main reason for the relatively low R values of the SVM model for P and I classification is the overlap of the word-frequency distributions of P and I sentences. By contrast, because of the use of L2 regularization and postpruning in XGBoost, the model prediction capabilities of the XGBoost model can be maximized while avoiding overfitting.

Of course, for our classification task, it is not truly meaningful to simply look at the P and R values of the models. We are more interested in evaluating model performance based on the F1 value, which is the harmonic mean of the P and R values. The F1 results indicate that the SVM model significantly outperforms the RF and XGBoost models. The ROC curves shown in Fig 3 also support this finding.

Table 8: Comparison of the soft-margin SVM model used in this study with the standard RF and XGBoost models.

|  | P elements | | | I elements | | | O elements | | |
| --- | --- | --- | --- | --- | --- | --- | --- | --- | --- |
|  | P | R | F1 | P | R | F1 | P | R | F1 |
| SVM | 0.925 | 0.838 | 0.879 | 0.842 | 0.789 | 0.814 | 0.886 | 0.897 | 0.891 |
| RF | 0.899 | 0.813 | 0.854 | 0.838 | 0.759 | 0.808 | 0.881 | 0.834 | 0.861 |
| XGBoost | 0.860 | 0.857 | 0.852 | 0.828 | 0.814 | 0.791 | 0.857 | 0.838 | 0.832 |

Note: The acc, P, R and F1 values obtained through 10-fold cross-validation are used to evaluate the models.

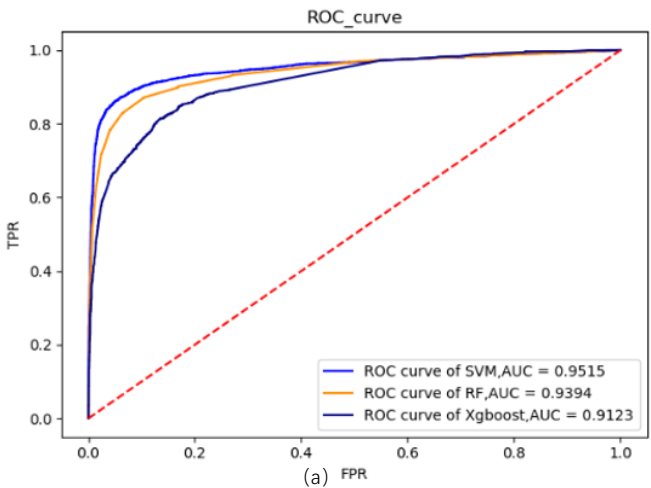

(a)

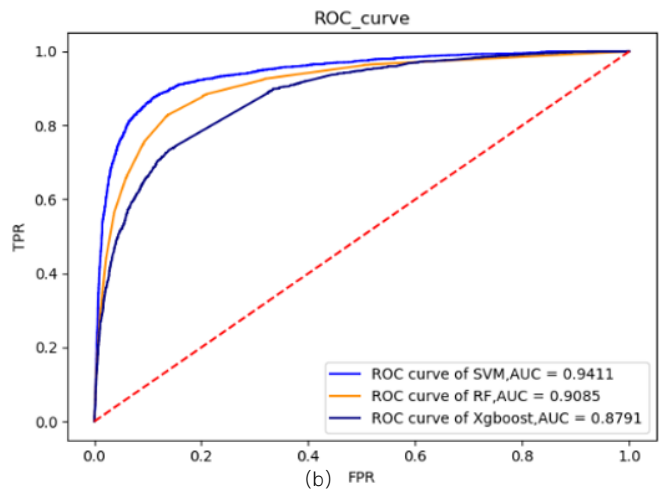

(b)

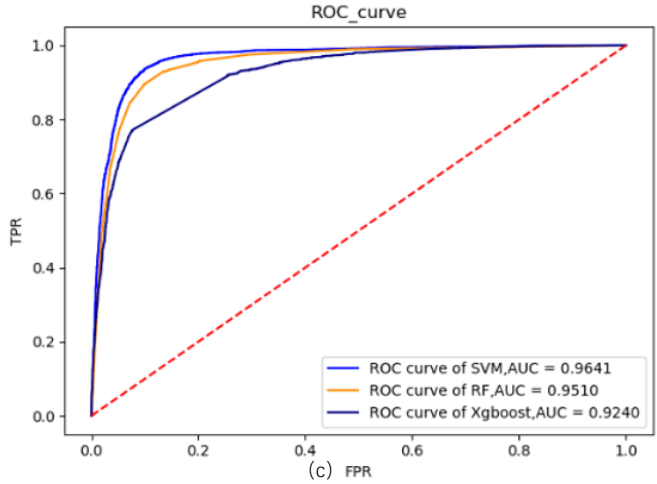

Fig 3: The ROC curves of the SVM model, RF model and XGBoost model comparison experiment: (a) results for P elements, (b) results for I elements, and (c) results for O elements.

## 5 Discussion

The data set used throughout this study is an open-source data set from BioNLP 2018 [11]. Because some previous studies have used different data sets and different methods, it is sometimes difficult to compare our findings with previous research results. However, since the problem studied is the same (PICO element extraction from RCT abstracts) and the model evaluation indicators are also the same (P, R, and F1 values), we can nevertheless compare the data presented in some previous research papers with the results of our model. In particular, since the data set used in this project is the same as the data set used in the Di Jin 2018 paper and is similar to the data set used in the Ke-Chun Huang 2011 paper, we will consider the results presented in those papers for comparison, as shown in Table 9.

Table 9: Comparison of model results.

|  | P elements | | | I elements | | | O elements | | |
| --- | --- | --- | --- | --- | --- | --- | --- | --- | --- |
|  | P | R | F1 | P | R | F1 | P | R | F1 |
| NB | 0.902 | 0.925 | 0.913 | 0.786 | 0.716 | 0.749 | 0.836 | 0.920 | 0.876 |
| LSTM | 0.885 | 0.828 | 0.856 | 0.749 | 0.815 | 0.781 | 0.845 | 0.832 | 0.838 |
| SVM | 0.925 | 0.838 | 0.879 | 0.842 | 0.789 | 0.814 | 0.886 | 0.897 | 0.891 |

Note: The LSTM model was described in the Di Jin 2018 paper, the NB model was described in the Ke-Chun Huang 2011 paper, and the SVM model is the model studied in the present paper.

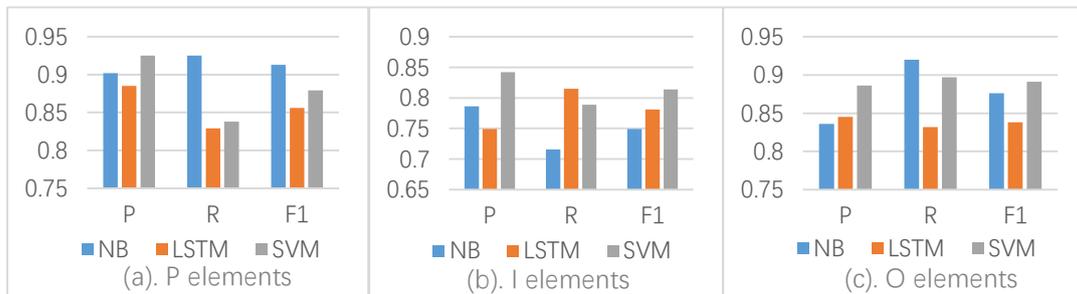

Fig 4: Histograms comparing the experimental results of the SVM model presented in this study, the LSTM model proposed by Di Jin 2018, and the NB model proposed by Ke-Chun Huang 2011: (a) results for P elements, (b) results for I elements, and (c) results for O elements. The three evaluation indicators, i.e., the P, R and F1 values, are represented on the horizontal axes.

In terms of accuracy (P value), the SVM model shows the best performance for P, I and O elements. In terms of recall (R value), the performance of the NB model is better than that of the SVM model for P and I elements, and the performance of the LSTM model is better than that of the SVM model for I elements. In terms of the F1 value, the SVM model shows the best performance for both I and O elements, but the performance of the SVM model for P elements is not as good as that of the NB model.

The reason for the differences evident in table 9 is that in the Ke-Chun Huang 2011 study, the authors used the words with the highest frequencies (1-14%) to represent the characteristics of each sentence, whereas we incorporated a 1-2gram model into our TF-IDF analysis to represent sentence characteristics. Using only the highest-frequency words to represent sentence features results in a considerable loss of information. By contrast, the TF-IDF approach not only retains the information contained in all words in a sentence but also highlights the characteristic information provided by low-frequency words that are unique to different sentences.

Regarding the introduction of the 1-2gram model, in the previous method and result sections, we have proven the superiority of the 1-2gram model for application to medical language through comparative experiments (see Table 5 for details). By contrast, the word2vec feature engineering method was used in the Di Jin 2018 study, and the results of this complex feature engineering method are not as good as those of our 1-2gram TF-IDF method, as can be seen by comparing the experimental results on the same data set.

As mentioned in the Di Jin 2018 paper, this method of constructing TF-IDF features based on probability statistics often ignores the correlations between different words and between different sentences. Therefore, its performance in natural language processing is not as good as that of word2vec. However, the TF-IDF method based on 1-2gram performs is much better than the word2vec method in terms of sentence feature representation, whether from the statistical results of word frequency of PICO sentences or the experimental results of TF-IDF comparison with word2vec. Because word2vec is better at capturing word similarity, word2vec is more suitable for context-based text understanding tasks. For our task, however, we believe that the feature representation method based on the word-frequency distribution is more beneficial for model training.

## 6. Conclusion

In this study, we have presented a TF-IDF feature engineering method that incorporates a 1-2gram model. We have also proposed a soft-margin SVM model based on multitask classification for the automatic extraction of sentence-level PICO elements from structured abstracts in the biomedical literature. We tested the performance of our model on an open-source data set and compared the results with those of previous research. We found that our model achieves the best F1 values for I and O elements. Moreover, although the F1 value of our model is slightly inferior to that of an NB model for P elements, our model is superior to the NB model in terms of accuracy (P value). Notably, in this study, we considered only the automatic extraction of PICO information from RCT abstracts. However, the PICO information will, of course, be more completely described in the full text. Therefore, in the future, we will use a deep learning method (self-attention) to construct more complex models for automatically identifying PICO information from the full text of RCT articles.

## 7.Acknowledgments

This project is funded by Sichuan Science and Technology Program Grant No.17PTDJ0078 & 182DYF0569. Thank Deqi Liu, the researcher of Sichuan Provincial Health and Family Planning Policy and Medical Information Research Institute, for his suggestions.